\documentclass[11pt,a4paper]{article}
\usepackage[hyperref]{ranlp2023}
\usepackage{times}
\usepackage{latexsym}

\usepackage{microtype}

\aclfinalcopy 

\title{Challenges and Applications of Automated Extraction of Socio-political Events from Text (CASE 2023): Workshop and Shared Task Report}

\author{Ali Hürriyetoğlu \\
  \small KNAW Humanities Cluster DHLab \\
  \texttt{\small name.surname@dh.huc.knaw.nl} \\\And
  Hristo Tanev \\
  \small European Commission \\
  \texttt{\small hristo.tanev@ec.europa.eu} \\\And
  Osman Mutlu \\
  \small Koc University \\
  \texttt{\small omutlu@ku.edu.tr} \\\AND
  Surendrabikram Thapa \\
  \small Virginia Tech \\
  \texttt{\small surendrabikram@vt.edu} \\\And
  Fiona Anting Tan \\
  \small National University of Singapore \\
  \texttt{\small tan.f@u.nus.edu} \\\And
  Erdem Yörük \\
  \small Koc University \\
  \texttt{\small eryoruk@ku.edu.tr}}

\date{}

\begin{document}
\maketitle
\begin{abstract}
We provide a summary of the sixth edition of the CASE workshop that is held in the scope of RANLP 2023. The workshop consists of regular papers, three keynotes, working papers of shared task participants, and shared task overview papers. 
This workshop series has been bringing together all aspects of event information collection across technical and social science fields. 
In addition to contributing to the progress in text based event extraction, the workshop provides a space for the organization of a multimodal event information collection task. 
\end{abstract}

\section{Introduction}

Nowadays, the unprecedented quantity of easily accessible data on social, political, and economic processes offers ground-breaking potential in guiding data-driven analysis in social and human sciences and in driving informed policy-making processes. Governments, multilateral organizations, and local and global NGOs present an increasing demand for high-quality information about a wide variety of events ranging from political violence, environmental catastrophes, and conflict, to international economic and health crises ~\cite{coleman2014handbook,della2015oxford} to prevent or resolve conflicts, provide relief for those that are afflicted, or improve the lives of and protect citizens in a variety of ways. The citizen actions against the COVID measures in the period 2020-2022 and the war between Russia and Ukraine are only two examples where we must understand, analyze, and improve the real-life situations using such data. Finally, these efforts respond to “growing public interest in up-to-date information on crowds” as well.~\footnote{\url{https://sites.google.com/view/crowdcountingconsortium/faqs}}

The workshop Challenges and Applications of Automated Extraction of Socio-political Events from Text (CASE 2023) is held in the scope of the conference Recent Advances in Natural Language Processing (RANLP). CASE 2023 is the sixth edition of a workshop series~\cite{hurriyetoglu-etal-2022-challenges,hurriyetoglu-etal-2021-challenges,hurriyetoglu-etal-2020-automated}.

We provide brief notes about the accepted papers, shared tasks, and keynote speeches in the following sections.

\section{Accepted papers}

This year, all seven submissions were accepted by the program committee. A quick summary of these papers are provided below.

\begin{itemize}
    \item ~\citet{1} collect and annotate a dataset in 3 different granularity: whether a document is related to criminal activities, whether each sentence is related to lethal behaviors, and each sentence's membership to 11 predefined categories of events. Following these granularities, the authors design three binary classification tasks and apply numerous non-neural and neural models to the annotated dataset; they observe good performance and provide analysis for data slices with lower performance.
    
    \item ~\citet{2} discern ``main" location, where the event in question occurred, versus ``secondary" locations, that provide extra context such as the origin of the protestors and the first event location in a series of protests. They accomplish this by training a BERT model on news articles annotated with the main event location (CASE 2021 shared task news dataset). The secondary locations are all other mentioned locations. They compared their results to that of two SVM models and a baseline that assumes only the first sentence contains the main location. Their model outperformed all other systems with an F1 of 0.8 and accuracy 0.73.

    \item ~\citet{3} proposes an extension of the Multiple-instance learning (MIL) framework to better handle a common problem in computational social science: given noisy reports from text sources, how can researchers identify if a bag of reports (here, tweets within a country-day) report a true event. The authors show that MIL improves civil unrest detection over methods based on simple aggregation. The experiments conducted on hyperparameters (key instance ratio n and instance supervision b) show an improvement from MIL-n compared to vanilla MIL and other variants by model selection. 

    \item ~\citet{7} This paper studies the news headlines event linking task that given a title of news (typically a sentence), maps it to an event concept from a knowledge graph. The challenge is how to compare different (zero-shot) models' performance. They propose a benchmark for the evaluation and compare multiple models. By comparing three families of approaches (a) similarity based on rule or embeddings, (b) off-the-shelf entity linking tools, and (c) prompting Large Language Models, the authors show that the approach (c) has the best performance even though different approaches could be complementary to each other.

    \item ~\citet{14} presents a semantic model to structure protest event ontology, and provides some general description of the practical work with the Bulgarian data. The paper presents both the modelling framework and the implementation. The model is a practical application of the Unified Eventity Representation (UER) formalism, which is based on the Unified Modeling Language (UML), whose four-layer architecture (i.e., user objects, model, metamodel, and meta-metamodel) provides flexible means for building the semantic representations of the language units along a scale of generality and specificity. 

    \item ~\citet{19} proposes a method that extends the detection capability of existing event detection models to new event types. The authors have an experimental setup for few-shot learning when there is a limited training sources. Moreover, they provide several analyses on the experimental results. The main strength is in using low resource (single GPU) in order to fine tune the model in a reasonable time by providing a small set of samples by leveraging the transfer learning feature of the pre-trained model. The performance on the detection of the already known events tends to improve as well.

    \item ~\citet{20} The paper propose a solution to address the issue of data scarcity in closed-domain event extraction. The proposed solution leverages on the use of a side-product of data annotation campaigns that are the data containing no annotation, by considering this information as a discriminant that improves the extraction performance. The authors propose a multi-task model where they leverage additional data present after the token annotation process. Experiments are well conducted, by showing the efficacy of the method using different gradually decreasing dataset dimensions.
    
\end{itemize}

\section{Shared tasks}

\subsection{Task 1: Multilingual Protest News Detection}


The performance of an automated system depends on the target event type as it may be broad or potentially the event trigger(s) can be ambiguous. 
The context of the trigger occurrence needs to be considered as well.
For instance, depending on the context, the ‘protest’ event type may or may not be synonymous with ‘demonstration’.
Moreover, the hypothetical cases such as future protest plans may need to be excluded from the results. 
Finally, the relevance of a protest depends on the actors, since only citizen-led events are in the scope of contentious political events.
This challenge becomes even harder in a cross-lingual and zero-shot setting where training data are not available in new languages.
We tackle the task in four steps and hope state-of-the-art approaches will yield optimal results.

This shared task was announced as a re-run of the same tasks from CASE 2021~\cite{hurriyetoglu-etal-2021-multilingual} and CASE 2022~\cite{hurriyetoglu-etal-2022-extended}. Although it attracted some interest, we did not receive any task description papers for this edition.

\subsection{Task 2: Automatically Replicating Manually Created Event Datasets}
The purpose of Task 2 is to test the abilities of the event extraction systems to map events on the World map, by finding the locations where they have taken place. This year the subtitle of the task was "Detecting and Geocoding Battle Events from Social Media Messages on the Russo-Ukrainian War": The purpose of the task was to detect armed clash events in Russian and Ukrainian Telegram messages and to geocode them, i.e., find the location of the battles at the level of a populated place~\cite{tanev2023geocoding}.

Until recently, event geocoding has been considered a topic, covered by the works in the area of Geoinformation systems. Only in the last years the NLP community started to consider ML algorithms for geocoding \cite{halterman2023mordecai}, also in the context of event detection \cite{tanev-where-2023-event}. In this context, we can consider our shared task as an evaluation exercise for such event geocoding systems.

The evaluation of the systems participating at shared task 2 relies on an original evaluation methodology which compares the battle coordinates, found by the systems with the locations of such events from a Gold standard data set. 

As a Gold standard this year we used a selected subset of the ACLED event database \cite{raleigh2010introducing}, covering the first six months start of the Russo - Ukrainian conflict, namely 24 February 2022 - 24 August 2022 and considering only the events of type {\em battle}. 

We provided the participants with English language text messages from Telegram channels originating from Russia and Ukraine. We gathered nearly 326K original English Telegram Massages from six Telegram channels.

The Telegarm data is available from 
a public Github repository \footnote{\url{https://github.com/htanev/RussoUkrainianWarTelegram}}.


Two systems participated in this year's evaluation: The top ranked system relied on a combination of a XLMRoberta \cite{liu2019roberta} classifier, trained on ACLED, and a rule based geocoder using the JRC NEROne named entity recognition system \cite{jacquet2019jrc}. 
The second ranked system used the NEXUS \cite{tanev2008real} keyword based event classifier and Mordecai3 geoparser \cite{halterman2023mordecai}.

The first system, based on XLMRoberta, achieved a moderate level of correlation with the ACLED dataset, which is a good result, considering the possibly low coverage of this year Telegram dataset, regarding Russo Ukrainian war.

We hypothesize that there is a low coverage of the Telegram dataset this year, since in the 2021 issue of this task \cite{giorgi2021discovering}, participating systems achieved much higher correlation with the ACLED Gold standard, including the NEXUS system.
The low correlation can be explained with the low coverage of the war, where battles are not always reported in the media, especially in Telegram. There is a lot of imprecise information on the social media and in contrast ACLED gold standard relies on a wide range of verified sources, including radio and TV, and manually curates the data.

Our conclusion from this year database replication task was that social media is not the best source of information about armed conflicts.


\subsection{Task 3: Event Causality Identification}

Causality is a core cognitive concept and appears in many natural language processing (NLP) works that aim to tackle inference and understanding. We are interested in studying event causality in the news and, therefore, introduce the Causal News Corpus \cite{tan-etal-2022-causal}. The dataset comprises of 3,767 event sentences extracted from protest event news, that have been annotated with sequence labels on whether it contains causal relations or not. Subsequently, causal sentences are also annotated with Cause, Effect and Signal spans. Two corresponding subtasks were involved in our shared task: In Subtask 1, participants were challenged to predict if a sentence contains a causal relation or not. In Subtask 2, participants were challenged to identify the Cause, Effect, and Signal spans given an input causal sentence. We hope that our shared task promotes research on the topic of detection and extraction of causal events in news.

This year's competition is the second iteration of the shared task, first introduced in 2022 \cite{tan-etal-2022-event}, and uses the latest version of the Causal News Corpus (CNC-V2), also known as RECESS \cite{tan-etal-2023-recess}.
As compared to V1 comprising of 160 sentences and 183 relations, the V2 contains 1,981 sentences and 2,754 causal relations for Subtask 2. Annotations were also revised for some examples across both subtasks. 

\citet{22} provides an overview of the work of the ten teams that submitted their results to our competition and the six system description papers that were received. The top F1 score for Subtask 1 was 84.66\% by Team DeepBlueAI, who did not submit a description paper. Team InterosML \cite{9} scored a similar high score of 84.36\%, and used a two step approach: first pre-training a baseline RoBERTa model with supervised contrastive loss, then fine-tuning the model on Subtask 1 itself. The top F1 score for Subtask 2 was 72.79\% by Team BoschAI \cite{6}, who used a sequence tagging approach to fine-tune BERT-large and RoBERTa-large, and adapted the target labels to allow prediction of up to three different causal relations per sentence.

\subsection{Task 4: Multimodal Hate Speech Event Detection}

Hate speech detection is one of the most important aspects of event identification during political events like invasions \cite{thapa2022multi}. In the case of hate speech event detection, the event is the occurrence of hate speech, the entity is the target of the hate speech, and the relationship is the connection between the two. Since multimodal content is widely prevalent across the internet, the detection of hate speech in text-embedded images is very important. 

Given a text-embedded image, task 4 aims to automatically identify the hate speech and its targets\footnote{\url{https://codalab.lisn.upsaclay.fr/competitions/13087}}. This task had two subtasks \cite{23}. In subtask 1, participants were given a dataset of text-embedded images and the participants had to classify whether the given image contained hate speech or not. It was tasked as a binary classification problem of classifying hate speech and non-hate speech. Similarly, in subtask 2, the participants were given a dataset of hateful text-embedded images where they had to classify what the targets of hate speech were. This subtask was posed as a multi-class classification problem where targets were individual, community, and organization. The dataset curated by \citet{bhandari2023crisishatemm} was used in this task. More than 50 participants registered for the competition.

\citet{23} presents the overview of the performance of 13 teams who submitted their scores in subtask 1 and 10 teams who submitted their scores in subtask 2. The ranking was done on the basis of the macro F1-score. The competition saw a wide range of methodologies ranging from traditional machine learning models to powerful transformer architectures. The first team ARC-NLP \cite{11} proposed an ensemble of multilayer perceptions (for representations from textual and visual encoders) and various boosting algorithms (using syntactical and Bag-of-words representations) for subtask 1. The team was able to score an F1-score of 85.65\%. Similarly, for subtask 2, they used Named Entity Recognition (NER) features along with CLIP representations. An ensemble approach similar to subtask 1 was able to give them the first position with an F1-score of 76.34\%. 

Similarly, many teams used transformer-based approaches. Out of the submitted papers for subtask 1, IIC\_Team \cite{21} ranked at rank 3 (F1-score 84.63\%), Ometeotl \cite{8} at rank 6 (F1-score of 80.97\%), and VerbaVisor \cite{12} at rank 8 (F1-score of 78.21\%) were able to get the best performances with XLM-Roberta-base, BertForSequence classification, and ALBERT models respectively. All of them used the text extracted from the given dataset of text-embedded images using Google Vision API. In subtask 2, IIC\_Team, VerbaVisor, and Ometeotl were able to get the rank of 3 with an F1-score of 69.73\%, rank 4 with an F1-score of 68.05\% and rank 7 with F1-score of 56.88\% respectively with same models used in subtask 1.

Often, the visual information is also necessary. The first team utilized both textual and visual information effectively to get a high F1-score. Two teams viz. CSECU-DSG and LexicalSquad leveraged both textual and visual information. CSECU-DSG \cite{17} used a combination of BERT and vision transformers (ViT) \cite{dosovitskiy2020image} to leverage textual and visual information respectively. They were able to get an F1-score of 82.48\% and 65.30\% in subtask 1 and subtask 2 respectively. They were placed at the fifth rank in both subtasks. Similarly, LexicalSquad \cite{13} participated only in subtask 1 where they used XLNet and BERT for textual features and Inception-V3 for visual features. With this combined representation, they were able to get an F1-score of 74.96\%. This ranked them at the tenth position in the leaderboard.

Traditional machine learning algorithms were also used by some teams where they performed decently well. SSN-NLP-ACE and ML\_Ensemblers used various traditional machine learning approaches. SSN-NLP-ACE \cite{10} used TF-IDF features with SVM (with RBF kernel) to get an F1-score of 78.80\% in subtask 1 ranking them in the seventh position in the leaderboard. They used TF-IDF features with logistic regression for subtask 2 which ranked them at eighth position with an F1-score of 52.58\%. Similarly, ML\_Ensemblers used a variety of algorithms like Naive Bayes, KNN, SVM, and Decision Trees out of which Naive Bayes performed the best in both subtasks with an F1-score of 42.94\% and 43.32\% in subtask 1 and subtask 2 respectively. They were able to secure the rank of 13 and 9 in subtask 1 and subtask 2 respectively.


\section{Keynotes}

Three scholars delivered three keynote speeches that are summarized below. 

\subsection{Using Automated Text Processing to Understand Social Movements and Human Behaviour}
\label{sec:eryoruk}

Erdem Yörük’s keynote will describe two large-scale ERC-funded projects that employs computational social science methods to extract data on protests and public opinion. The first is the Global Contentious Politics Dataset (GLOCON) Project.~\footnote{\url{https://glocon.ku.edu.tr/}} is the first automated comparative protest event database on emerging markets using local news sources~\cite{durusan2022global}. The countries included in the GLOCON dataset are India, South Africa, Argentina, Brazil and Turkey. Glocon has been created by using natural language processing, and machine learning in order to extract protest data from online news sources. The project develops fully automated tools for document classification, sentence classification, and detailed protest event information extraction that performs in a multi-source, multi-context protest event setting with consistent performances of recall and precision for each country context. GLOCON counts the number of events such as strikes, rallies, boycotts, protests, riots, and demonstrations, i.e. the “repertoire of contention,” and operationalizes protest events by various social groups. The project has developed a novel bottom-up methodology that is based on a random sampling of news archives, as opposed to keyword filtering. The high-quality Gold standard corpus is designed in a way that can accommodate context variability from the outset as it is compiled randomly from a variety of news sources from different countries~\cite{hurriyetoglu-et-al-2021-cross-context,yoruk-et-al-2021-randomly}. The second one, Politus Project, aims at scaling up traditional survey polls for public opinion research with AI-based social data analytics. Politus develops an AI-based innovation that combines quantitative and computational methods to create a data platform that delivers representative, valid, instant, real-time, multi-country, and multi-language panel data on key political and social trends. The project will collect content information from Twitter and process it with AI tools to generate a large set of indicators on political and social trends through its data platform. The deep learning models and NLP tools will be designed from the ground up as language-independent and generalizable systems. The platform will deliver geolocated hourly panel data on demography, ideology, topics, values, and beliefs, behavior, sentiment, emotion, attitudes, and stance of users aggregated at the district level. In this keynote, Dr. Yörük will describe the general methodology of the projects, including data collection, data analysis, and their approach for representativeness, which is based on multilevel regression with post-stratification.

\subsection{Bulgarian Event Corpus for the Construction of a Bulgaria-centric Knowledge Graph}
\label{sec:kisimov}

The Bulgarian Event Corpus is being constructed within the CLaDA-BG (Bulgarian National Interdisciplinary Research E-Infrastructure for Bulgarian Language and Cultural Heritage Resources and Technologies. In the spirit of European CLARIN and DARIAH) we aim to support researchers in Humanities and Social Sciences (H\&SS) to access the necessary datasets for their research. The different types of objects of study, representation and search are integrated on the basis of common metadata and content categories. The approach for interlinking of the datasets is called contextualization. The implementation of contextualization in CLaDA-BG will utilize a common Bulgaria-centered knowledge graph - BGKG. The knowledge facts within BGKG are constructed around events of different types. Thus, construction of BGKG requires a set of appropriate language resources for training of Bulgarian language pipeline for extraction of events from text documents. A key element within these language resources is the Bulgarian Event Corpus.
In the talk I will present the design of the annotation schema, the annotation process, relation to ontologies and RDF representation. We have started with the CIDOC-CRM ontology for the construction of the annotation schema. This ontology provides a good conceptualization of events motivated by the domain of museums which is appropriate for our goals. During the design of the annotation schema, we extended the ontology with new events depending on the content of the corpus. The documents to be annotated were selected from scientific and popular publications of the partners within CLaDA-BG and articles from Bulgarian Wikipedia. The annotation is done on several layers: Named Entities, Events, Roles, Linking, terms and keywords.

\subsection{With a little help from NLP: My Language Technology applications with impact on society}
\label{sec:rumitkov}

Ruslan Mitkov will present original methodologies developed by the speaker, underpinning implemented Language Technology tools which are already having an impact on the following areas of society: e-learning, translation and interpreting and care for people with language disabilities.

The first part of the presentation will introduce an original methodology and a tool for generating multiple-choice tests from electronic textbooks. The application draws on a variety of Natural Language Processing (NLP) techniques which include term extraction, semantic computing and sentence transformation. The presentation will include an evaluation of the tool which demonstrates that generation of multiple-choice tests items with the help of this tool is almost four times faster than manual construction and the quality of the test items is not compromised. This application benefits e-learning users (both teachers and students) and is an example of how NLP can have a positive societal impact, in which the speaker passionately believes. The latest version of the system based on deep learning techniques will also be briefly introduced.

The talk will go on to discuss two other original recent projects which are also related to the application of NLP beyond academia. First, a project, whose objective is to develop next-generation translation memory tools for translators and, in the near future, for interpreters, will be briefly presented. Finally, a project will be outlined which focuses on helping users with autism to read and better understand texts. The speaker will put forward ideas as to what we can do next.

The presentation will finish with a brief outline of the latest (and forthcoming) research topics (to be) which the speaker plans to pursue and his vision on the future NLP applications. In particular, he will share his views as to how NLP will develop and what should be done for NLP to be more successful, more inclusive and more ethical.

\section{Conclusion}

Many aspects of event information modeling and collection are reported in the scope of CASE 2023. Hosting a shared task that is on multimodal problem and having submissions about languages other than English (e.g., Bulgarian) are distinguishing aspects of this edition.

\bibliographystyle{acl_natbib}
\bibliography{ranlp2023}


\end{document}